\documentclass[letterpaper, 10 pt, conference]{ieeeconf}  

\IEEEoverridecommandlockouts                              

\usepackage{graphicx}

\usepackage{amsmath} 
\usepackage{amssymb}  

\usepackage{caption}
\usepackage{subcaption}

\usepackage{cite}
\usepackage{amsfonts}
\usepackage{hyperref}

\usepackage{booktabs, makecell}
\usepackage{threeparttable}

\usepackage{multirow}

\newtheorem{remark}{Remark}

\title{\LARGE \bf
Contour Context: Abstract Structural Distribution for 3D LiDAR Loop Detection and Metric Pose Estimation
}

\author{Binqian Jiang and Shaojie Shen
\thanks{This work was supported by HKUST Postgraduate Studentship and HKUST-DJI Joint Innovation Laboratory.
	The authors are with the Department of Electronic and Computer Engineering, The Hong Kong University of Science and Technology, Hong Kong, China. E-mail: 
        {\tt\small bjiangah@connect.ust.hk}, {\tt\small eeshaojie@ust.hk}}%
}

\begin{document}

\maketitle
\thispagestyle{empty}
\pagestyle{empty}

\begin{abstract}
	
This paper proposes \textit{Contour Context}, a simple, effective, and efficient topological loop closure detection pipeline with accurate 3-DoF metric pose estimation, targeting the urban autonomous driving scenario. We interpret the Cartesian birds' eye view (BEV) image projected from 3D LiDAR points as layered distribution of structures. 
To recover elevation information from BEVs, we slice them at different heights, and connected pixels at each level will form contours. Each contour is parameterized by abstract information, e.g., pixel count, center position, covariance, and mean height. 
The similarity of two BEVs is calculated in sequential discrete and continuous steps. The first step considers the geometric consensus of graph-like constellations formed by contours in particular localities. 
The second step models the majority of contours as a 2.5D Gaussian mixture model, which is used to calculate correlation and optimize relative transform in continuous space.
A retrieval key is designed to accelerate the search of a database indexed by layered KD-trees. 
We validate the efficacy of our method by comparing it with recent works on public datasets. 

\end{abstract}

\section{Introduction}  \label{sec:intro}
Towards the goal of robust and long-term autonomy, a fundamental task in perception is to recognize previously visited places. This task is found in many practical and critical robotic applications, e.g., wake-up on a pre-built map, recovery from localization failures, and simultaneous localization and mapping (SLAM) \cite{segmatch}.
Much research has been devoted to vision-based loop detection \cite{orbslam, Lowry_Sunderhauf_Newman_Leonard_Cox_Corke_Milford_2016}, but their methods often suffer from visual feature variation due to illumination, view point or seasonal changes. Light detection and ranging (LiDAR) sensors are more resilient to appearance changes by accurately capturing the underlying structural information\cite{scan_context, pointnetvlad, overlapnet, Bosse2013, segmap_conf}, suggesting valuable potential for global localization. 


In autonomous driving, the relative pose of road vehicles typically varies little in roll, pitch, and vertical direction when revisits occur, so it is reasonable to assume that cars' pose lives in 2D when relocalizing. This assumption encourages many \cite{scan_context, scan_context_pp, bvmatch, disco_lc, lidar_iris, spi_lc} to use birds' eye view (BEV) projections of 3D point clouds for loop detection. 
However, how to efficiently generate effective place descriptors that emphasize the invariance of scenes remains an open issue. Polar projection based methods typically need to shift descriptors when looking for the best alignment, which is essentially brutal force search\cite{scan_context, scan_context_pp,lidar_iris}. Also, they are sensitive to relative translation. 
Image feature detectors and descriptors rely on the invariance of intensity (or its gradient change) in a small area. While in 3D LiDAR Cartesian BEV, this can be unstable due to the sparsity of LiDAR points and sensor motion. 
To mitigate this problem, Cartesian projection based methods often follow the image matching pipeline, which focuses on image filtering and key points extraction and matching\cite{bvmatch, spi_lc}. They are resource demanding, since a bank of filters are used, and too many points are involved in relative pose estimation. 


To better use BEV images in loop detection, 
we propose \textit{Contour Context} (Cont2), a simple, effective, and efficient topological loop closure detection pipeline with accurate 3-DoF metric pose estimation (MPE). 
Our key insight is that the scene expressed in BEV can be modeled as a probability distribution of salient structures. Distributions are more stable and repeatable than points since they are an ``average" in general. 
This new interpretation exploits the embedded structure information unique to LiDAR BEVs, which is seldom touched in previous research.
By keeping the highest point projected into each BEV pixel, we preserve map elevation. Slicing the BEV image at different heights creates contours (i.e. connected pixels) at each level, which contain 2D structural information. 
We compress a raw contour into an ellipse and a few other variables, a compact and uniform parameterization of the environment components. 
The similarity of two BEVs is calculated in sequential discrete and continuous steps. The first step checks the consensus of graph-like constellations formed by multiple contours in a locality.
The second step models the scene as a 2.5D Gaussian mixture model (GMM) to calculate correlation and optimize the transform parameters in continuous space.
A retrieval key characterizing the vicinity of some contours allows for fast retrieval of promising candidates. Our method is competitive compared with many recent works on different datasets.
The main results of our research reported in this paper include:
\begin{itemize}

	\item A novel interpretation of 3D LiDAR BEV images as layered distribution of structures.
	\item A two-step similarity check using discrete constellation consensus verification and L2 optimization with continuous densities. 

	\item An effective and real-time topological loop closure detection pipeline with accurate 3-DoF metric pose estimation. We will release our source code for the readers' reference\footnote{https://www.github.com/lewisjiang/contour-context}. 


\end{itemize}

\section{Related Work} \label{sec:related_work}
\subsection{General Global Descriptor Based Methods}
Place recognition on 3D point clouds can be solved by extracting 3D key points and using them to describe the scene. In \cite{Bosse2013}, key points are extracted directly in 3D, and the votes cast by these points are used to determine the loop candidates. PointNetVLAD\cite{pointnetvlad} and LCDNet\cite{lcdnet} extracts features using DNN-based point feature extraction.

Global descriptors can be generated without explicit feature points. M2DP\cite{m2dp} uses point distribution of different planes as global descriptor. GLARE\cite{glare_2d} maps relative azimuth angle and distance of point pairs into a global description matrix. OverlapTransformer\cite{OverlapTransformer}, which generalizes OverlapNet\cite{overlapnet}, creates global descriptors from range images using depth only clues. While one global descriptor per scan is common, our method generates descriptors from part of the input data, and one scan can have multiple descriptors.

\subsection{BEV Based Methods}
Many methods adopt polar BEV projection, assuming a small translation difference when revisits occur. ScanContext\cite{scan_context} maps the radial and azimuthal coordinate of 3D points into matrix row and column, and use ring features for retrieval. DiSCO\cite{disco_lc} transforms polar BEV into the frequency domain, making the descriptor yaw-invariant. LiDAR Iris\cite{lidar_iris} uses LoG-Gabor filters on polar BEV image strips to increase the representation ability. 

Another type of BEV projection uses Cartesian coordinates in 2D. ScanContext++\cite{scan_context_pp} augments matrix descriptors by shifting and sequential flips. Other methods mainly use image matching pipelines, though stable image features are difficult to extract. Using image feature detectors and descriptors to match 2D LiDAR occupancy grids is studied in \cite{Blanco2013}. SPI\cite{spi_lc} uses a NetVLAD global descriptor on submap BEV and SuperPoint and SuperGlue for feature matching. BVMatch\cite{bvmatch} extracts features with FAST detector after applying Log-Gabor filters to the BEV image. These methods focus on finding invariance on pure images, which is intuitive but neglects the characteristics of 3D LiDAR point clouds.

\subsection{Object and Graph Based Methods}
Object and segment-based methods gradually gain popularity in recent years. SegMatch\cite{segmatch} extracts segments from local point cloud map and describe them using handcrafted scores. SegMap\cite{segmap_conf, segmap_j} improves SegMatch with data-driven descriptors, and an efficient geometrical consistency verification is used in \cite{Dube2018}.
Object-based methods are closely related to graphs.
SGPR\cite{sg_pr} uses semantic segmentation for a graph of instances, and a graph similarity network for matching. GOSMatch\cite{gosmatch} employs only 3 kinds of semantic categories with 6 kinds of pairwise relations, and it uses histogram-based vertex and graph descriptors for matching. Seed\cite{Fan_He_Tan_2020} segments point clouds into objects and puts the polar BEV coordinate at a primary object before object projection. BoxGraph\cite{boxgraph} uses the bounding box of objects as graph vertex and quantifies the pairwise similarity of vertices and edges to find optimal graph matching. Locus\cite{locus_lc} encodes structural-appearance features of segments and their repeatability across frames. Segments and objects are a compact representation of the environment, but they rely on segmenters or semantic detectors, which may be brittle due to domain gaps in the real world. Also graph-based methods typically need pairwise similarity check\cite{sg_pr, boxgraph}, it's unclear how to do efficient retrieval.

\section{Contours on BEV}

\begin{figure}[tpb]

	\centerline{\includegraphics[width=0.4\textwidth]{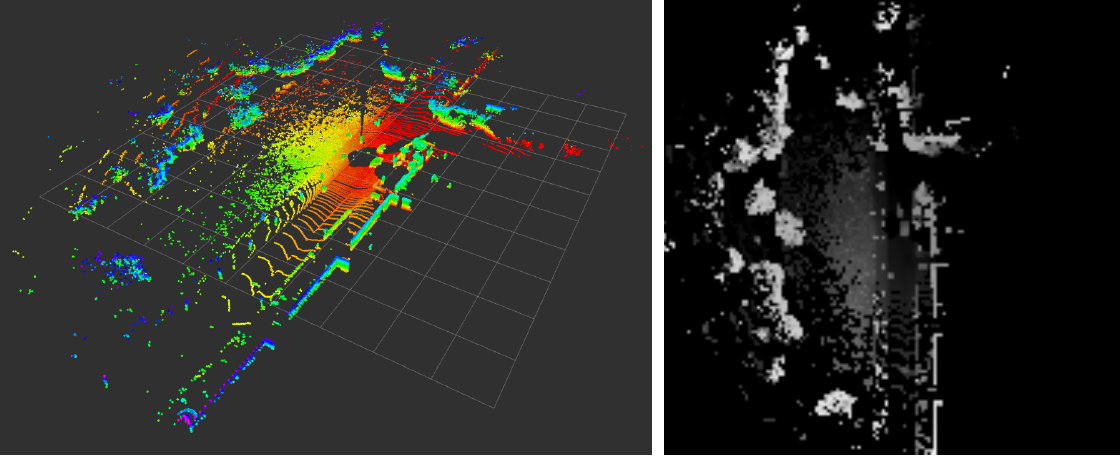}}

	\caption{A sample point cloud and its Cartesian BEV. Data from laser scan No.1648, KITTI odometry sequence 08. }

	\label{fig:pc_bev}

\end{figure}

\begin{figure*}[htbp]

	\centering 

	\includegraphics[width=0.9\textwidth]{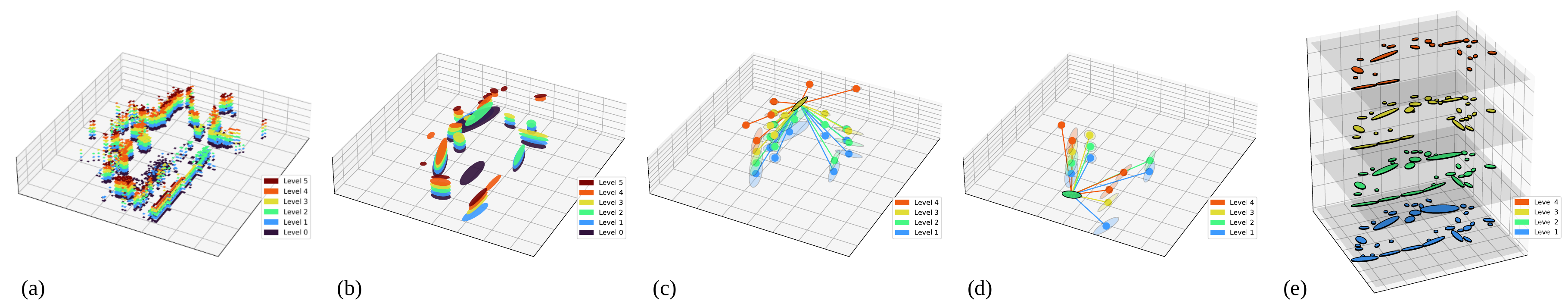}

	\caption[]{(a) A sliced BEV image. (b) Top-10 CAs at each level. (c)(d) Two CACs at different anchors. Ellipses with solid colors (yellow in (c), green in (d)) and black edges are anchors, others are peripheral. (e) CAs used as 2.5D GMM components.}

	\label{fig:sim_res}

\end{figure*}

In this section, we introduce the concept of \textit{contour abstraction} (CA). 
Our projection step is inspired by the \textit{Cart Context} in ScanContext++ \cite{scan_context_pp}, which uses the Cartesian coordinate to create image-pixel-like bins in 2D and records the maximum height of points in each bin (Fig. \ref{fig:pc_bev}). 
For economy of space, we refer readers to \cite{scan_context_pp} for a detailed definition.
For later reference, the notation of a pixel $\mathbf{p}$ is
\begin{equation}
	 \mathbf{p} = [p_x, p_y, p_z]^{\top}, \quad \mathbf{p}_{xy} = [p_x, p_y]^{\top}
\end{equation}
where $\mathbf{p}_{xy}$ is the 2D coordinate of a pixel in the BEV coordinate frame and $p_z$ is its height (i.e., pixel value).


A BEV image is sliced at preset heights to give vertical structural information (Fig. \ref{fig:sim_res}a). 
The pixels at each level that are 8-connected form a contour, hence the name \textit{Contour} Context. The height discretization paves way for layer-wise operations and description of isolated contours.


Focusing on statistic quantities, we represent individual contours with 2D Gaussian distributions (Fig. \ref{fig:sim_res}b) and some descriptive variables. This aggravates information loss, but it enables us to encode the contours of various shapes with a uniform parameterization. 
We define a \textit{raw contour} as a collection of 8-connected pixels, and $\mathcal{A}^{l, s}$ denotes the $s$-th raw contour at level $l$ sliced off a BEV image. The complete parameterization of a contour abstraction $\mathcal{C}^{l, s}$ of $\mathcal{A}^{l, s}$ is
\begin{equation} \label{eq:def_abs_cont}
\mathcal{C}^{l, s} = \{ n_{a}, h_{m}, \mathbf{x}_c, \mathbf{x}_{m}, \mathbf{C}, \mathbf{v}_{1},\mathbf{v}_{2}, {\lambda}_{1}, {\lambda}_{2} \},
\end{equation}
where
\begin{equation}
\label{eq:def_explain}
\begin{aligned}
	n_{a} &= | \mathcal{A}^{l, s} |, \quad h_m = \frac{1}{n_a} \sum_{\mathbf{p} \in \mathcal{A}^{l, s}}{p_z}, \\
	\mathbf{x}_c &= \frac{1}{n_{a}} \sum_{\mathbf{p} \in \mathcal{A}^{l, s}}{ \mathbf{p}_{xy} }, \quad
	\mathbf{x}_m = \frac{1}{n_{a}} \sum_{\mathbf{p} \in \mathcal{A}^{l, s}}{ p_z \cdot \mathbf{p}_{xy} }, \\
	\mathbf{C} &=  \frac{1}{{n_{a}-1}} \sum_{\mathbf{p} \in \mathcal{A}^{l, s}}{(\mathbf{p}_{xy}-\mathbf{x}_{c})(\mathbf{p}_{xy}-\mathbf{x}_{c})^{\top} }, 
\end{aligned}
\end{equation}
and the physical interpretations: $n_{a}$ is the number of pixels, $h_m$ is the mean height, $\mathbf{x}_c$ is the center of mass (CoM), $\mathbf{x}_m$ is height-weighted CoM, and $\{\mathbf{v}_i,\lambda_i\}$ is an eigenvector-eigenvalue pair of shape covariance $\mathbf{C}$ with $\lambda_{1}\ge\lambda_{2}$.

Big CAs summarize tens of pixels from the BEV image, which makes them remarkably reliable for preserving the \textit{mean position} (i.e., $\mathbf{x}_c$) of contours in 2D. To differentiate the importance of CAs at each level,  we sort them in descending order of $n_a$ and then index them with sequence id $s$.


\section{Loop Detector Design} \label{sec:design}

\begin{figure}[bp]
	
	\centerline{\includegraphics[width=0.45\textwidth]{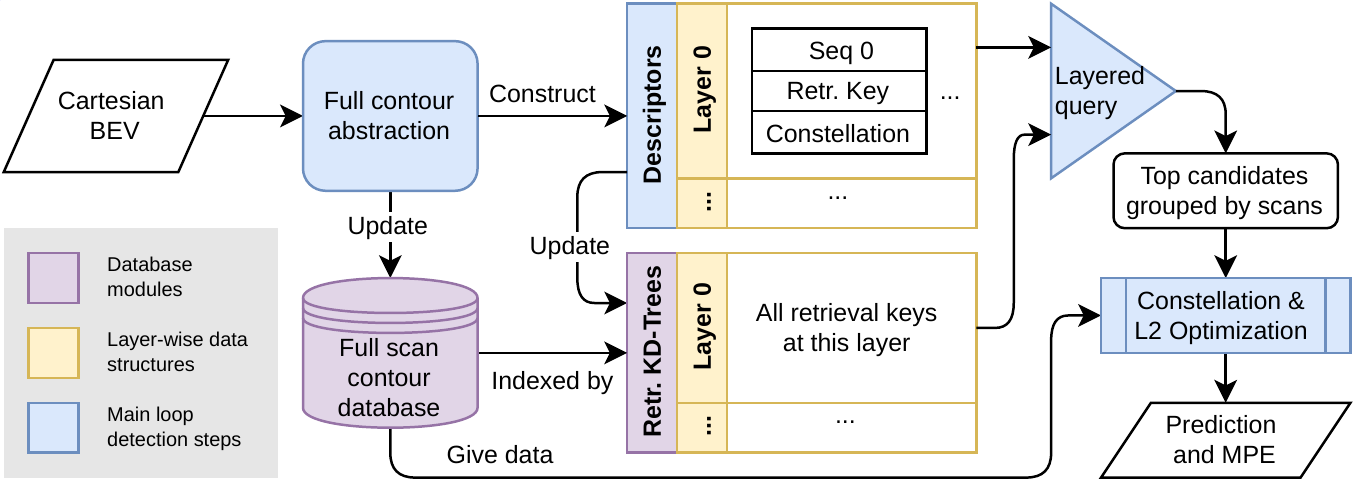}}

	\caption{The pipeline of the proposed loop detection system. }
	
	\label{fig:diagram}
	
\end{figure}

\subsection{Constellation of Contours} \label{sec:relational}


We use a graph-like structure called the \textit{contour abstraction constellation} (CAC) for partial BEV description. It consists of an anchor CA and several nearby peripheral CAs at various levels (Fig. \ref{fig:sim_res}c, \ref{fig:sim_res}d). 
We use three sequential steps to check for the similarity of two CACs, which are candidates from retrieval (Sec. \ref{sec:retr}, Fig. \ref{fig:diagram}). 

\subsubsection{Anchor CA similarity} Two CAs going for a similarity check should come from the same level. We reject the CA pair if selected parameter pairs do not agree. We define a similarity check function $\operatorname{SC}$ with binary output for scalars:
\begin{equation} \label{eq:sim_check}
\operatorname{SC}_{t_a, t_p}(x_1, x_2)= 
\begin{cases}
1, & \frac{|x_1-x_2|}{\max{(x_1, x_2)}}<t_p \text{ or } |x_1-x_2|<t_a;\\
0, & \text{otherwise.}
\end{cases}    
\end{equation}
where $x_1$ and $x_2$ are the two variables in a check, with $t_p$ and $t_a$ as the percentage and absolute difference thresholds. There are five scalars going through the test ($ n_{a}, h_{m}, \|\mathbf{x}_c-\mathbf{x}_{m}\|_{2}, {\lambda}_{1} \text{  and  } {\lambda}_{2}$), and each uses its own set of $t_p$ and $t_a$.

\subsubsection{Constellation structure similarity} This step looks for a relative transform with the consensus of as many peripheral CA CoM pairs as possible. With the similarity of anchor CA pair satisfied, their CoMs are assumed to be identical, then only a rotation in 2D is left to be found. To tackle this problem efficiently, we divide it into two smaller problems: find the CAs with roughly the same distance to the anchor and then find the rotation voted by most peripheral CAs.

In the first minor step, we use a binary vector $\mathbf{b}$ to record the coarse distance and level of peripheral CAs, which is a zero vector with some bits set as: 
\begin{equation}
\mathbf{b}(\phi(d, l)) = 1
\end{equation}
where $\phi(d, l)$ maps the peripheral-anchor distance $d$ and peripheral level $l$ to bits. It is calculated during scan preprocessing. Using bit-wise $\operatorname{AND}$ of two binary vectors, we can establish potential peripheral CA pairs with similar distance-to-anchor (result of $\operatorname{AND}$). The list of candidates is
\begin{equation}
L_{dist} = [\{ {}^{1}_{1}\mathcal{C}, {}^{2}_{1}\mathcal{C}\}, \cdots, \{ {}^{1}_{i}\mathcal{C}, {}^{2}_{i}\mathcal{C}\}, \cdots]
\end{equation}
where $\{ {}^{1}_{i}\mathcal{C}, {}^{2}_{i}\mathcal{C}\}$ comprises of paired peripheral CAs from $1$-st and $2$-nd CAC respectively. 

As for the second minor step, we sort $L_{dist}$ according to $\psi( {}^{1}_{i}\mathcal{C}) - \psi({}^{2}_{i} \mathcal{C})$, where $\psi({}^{j}_{i}\mathcal{C})$ calculates the azimuth angle of the peripheral CA measured at its CAC anchor. Comparing azimuth angles from different frames is normally meaningless since they are measured in local coordinates. However, if two peripheral CAs are a true match, taking the difference between the two azimuth angles will reveal the relative rotation between the two local coordinates plus some constant offset. 
In short, the CA pairs voting for the same rotation have unknown but close $\psi( {}^{1}_{i}\mathcal{C}) - \psi({}^{2}_{i} \mathcal{C})$. We can find the angle clustering with the most votes by iterating over the sorted $L_{dist}$ with some angle window in just one pass. The constellation is actually fixed with both distance and rotation pinned down in the CAC anchor frame. 

\subsubsection{Constellation pairwise similarity} \label{sec:constell}
Using the same similarity check function in (\ref{eq:sim_check}), we further inspect the paired peripheral CAs from the previous step. 
By adjusting the pair number threshold for remaining CA pairs, we can throttle the number of positive predictions fed into next stage.



\subsection{GMM-based Distribution Modeling} \label{sec:gmm}
\begin{figure}[b]
	
	\centerline{\includegraphics[width=0.3\textwidth]{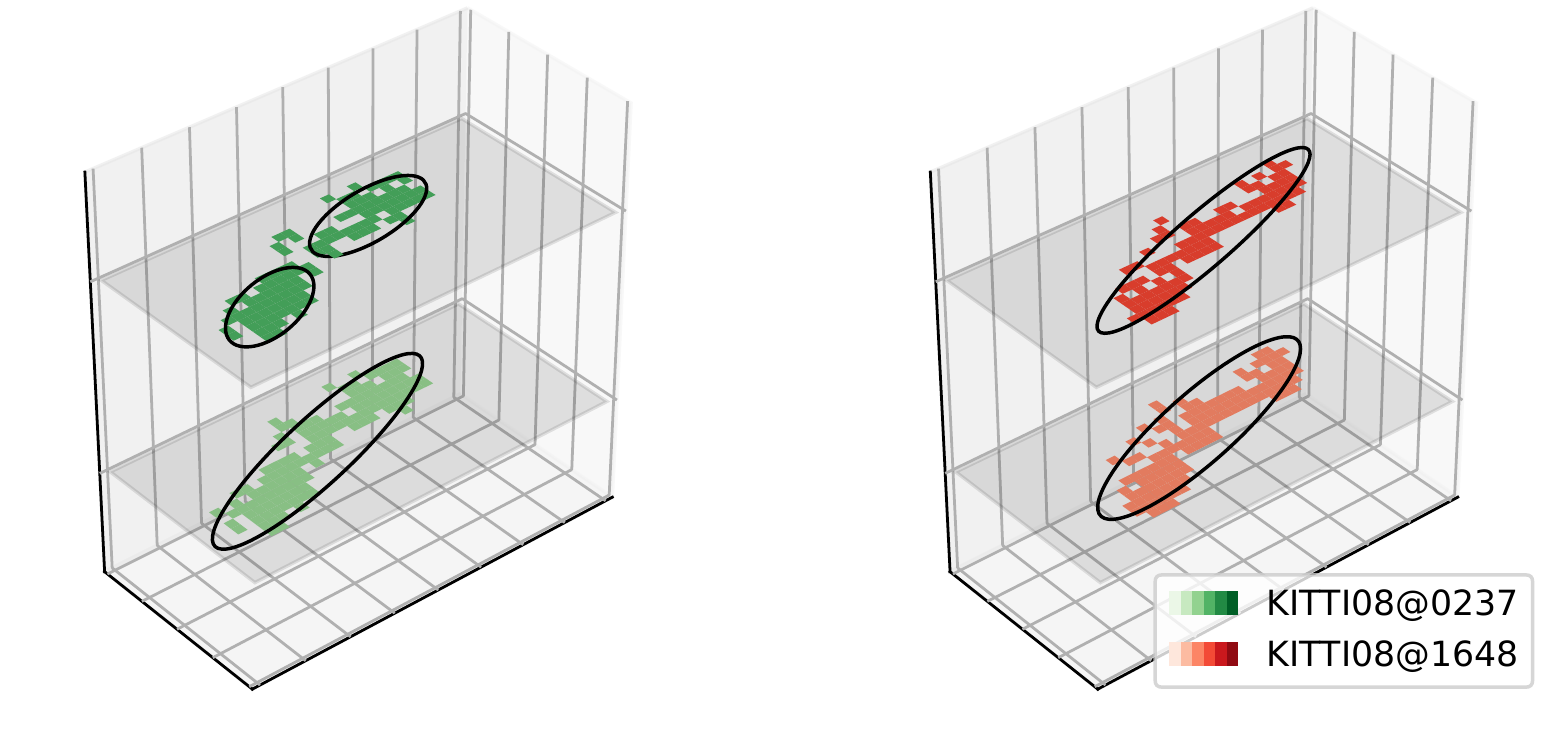}}

	\caption{Illustration of inconsistent discretization. Similar CAs are no longer similar when thresholding at a higher level. The black ellipses are the 2$\sigma$ boundaries of the underlying CAs.}
	
	\label{fig:conhesion}
	
\end{figure}


\subsubsection{Intuition and formulation}
The estimated transform from Sec. \ref{sec:constell} can be less accurate or even wrong, as many CAs are wasted due to the requirement for pairwise CA similarity. 
This is usually caused by inconsistent discretization results. For example in Fig. \ref{fig:conhesion}, intuitively a combination of the two CAs on the upper left should be highly similar to the single CA of the same level on the right. 
To quantify this idea,
we further model the similarity check as a 2.5D distribution-to-distribution (D2D) correlation maximization problem. 
It enlarges the region of attraction compared to those with deterministic correspondences. 
We first define the GMM constructed from CAs of a scan in 3D:
\begin{equation}  \label{eq:gmm_3d}
\begin{aligned}
gmm_{C}(x,y,z) = \frac{1}{N_a} \sum_{l \in L} & \sum_{s} n^{l, s}_{a} \delta(z-l) \cdot \\
& \mathcal{N}([x,y]^\top \mid \mathbf{x}^{l, s}_{c}, \mathbf{C}^{l,s})
\end{aligned}
\end{equation}
where $L$ is the set of levels of interest, $\delta(\cdot)$ is the Dirac delta function, $\mathcal{N}\left({x} \mid \mu, \Sigma\right)$ is a Gaussian PDF with mean $\mu$ and covariance $\Sigma$ evaluated at $x$, and $N_a = \sum_{l \in L} \sum_{s} n^{l, s}_{a}$. The definition in (\ref{eq:gmm_3d}) is a probability density function in $\mathbb{R}^3$ (i.e., integral over $\mathbb{R}^3$ is exactly $1$). An illustration of the mixture components (CA ellipses) is shown in Fig. \ref{fig:sim_res}e. CAs at  less informative levels are excluded. The mixture components are weighted by their pixel area, which is the only clue for which mixture a random pixel belongs to.

As a BEV-based approach, our transform estimate only includes $x$, $y$, and yaw angle. In addition, the third dimension \textit{level} is not comparable with the other two. It is added together with the Dirac delta, so layer-wise operations and optimization in continuous space can be precisely defined. Therefore, we call this approach 2.5D-based.

\subsubsection{Optimization}
To measure the similarity between two GMMs, we adopt L2 distance, which is a special case of a more general divergence family and has closed-form expression for Gaussian\cite{gmmreg}. The general form of L2 distance between two distributions $f$ and transformed $g$ is written as
\begin{align}  
\label{eq:L2}  
d_{L_2}(f, g, \theta)&=\int(f-T(g, \theta))^2 d x \\
\label{eq:expand}
d_{L_2}(f, g, \theta)&=\int f^2 d x - 2\int f T(g,\theta) d x + \int g^2 d x
\end{align}
where $T(\cdot,\theta)$ is an isometric transform with parameter $\theta$. Since $T$ is isometric and the interval of integration is the whole space, $\int g^2 d x = \int T^2(g, \theta) d x$. The first and third terms on the right side of (\ref{eq:expand}) can be calculated without any knowledge of $\theta$. We can express probability density correlation $\operatorname{cor}(\cdot,\cdot)$ using the same variable terms in (\ref{eq:expand}) as:
\begin{equation}
\operatorname{cor}(f, h) = \frac{\int f h d x}{\sqrt{\int f^2 d x \int h^2 d x}}, \; \text{with}\; h = T(g, \theta)
\end{equation}
Obviously, minimizing the L2 distance is equivalent to maximizing the correlation between two densities, while correlation is a normalized metric. Finally, we can formulate the transform parameter and correlation optimization problem as:
\begin{equation}  \label{eq:gmm_argmin}
\hat{\theta} = \arg \min_{\theta}{ \int - gmm_{C_1}(\mathbf{x}) \cdot T\left(gmm_{C_2}(\mathbf{x}), \theta \right) d \mathbf{x}}
\end{equation}
where $ gmm_{C_1}(\mathbf{x})$ and $ gmm_{C_2}(\mathbf{x})$ are 2.5D GMMs of two scans as defined in (\ref{eq:gmm_3d}), $\mathbf{x}$ is the vectorized variable, and $\theta$ is the transform parameter limited to $x$, $y$ and yaw only. We use the optimized correlation as the final similarity score.


\begin{remark}
	
	We use Ceres\cite{ceres} to solve the general unconstrained minimization problem in (\ref{eq:gmm_argmin}) with analytic derivatives. A similar minimization reported in 3D NDT D2D registration\cite{l2_3d_ndt} does not take the derivative of denominators, which actually can not minimize the objective function.
\end{remark}

\begin{remark}
	In practice, we ignore CA pairs that are too far apart in exchange for efficiency during optimization. This will cause a permanent underestimate of the current correlation value. However, the trade-off is worthwhile because the contribution of the pruned terms is insignificant.
\end{remark}
\subsection{Retrieval Design}  \label{sec:retr}

\begin{figure}[tbp]
	
	\centerline{\includegraphics[width=0.4\textwidth]{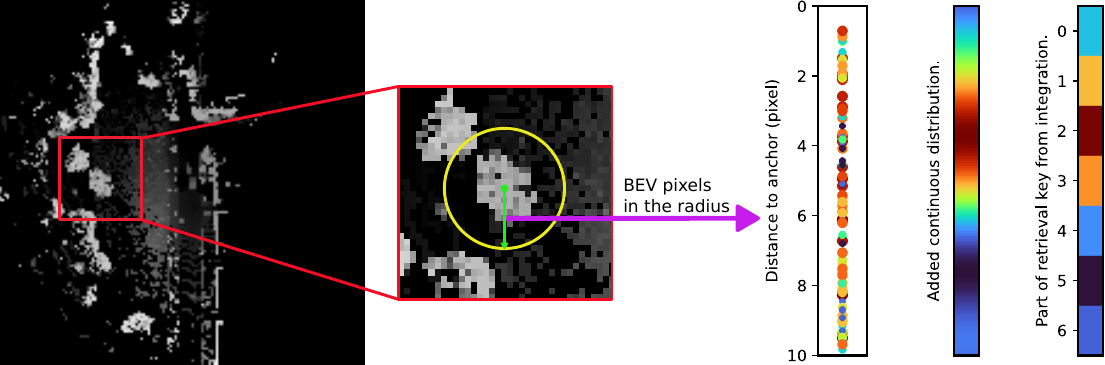}}

	\caption{Illustration of the dimensions of retrieval key generated from the vicinity (yellow circle) of an anchor CA. }
	
	\label{fig:gen_ret_key}
	
\end{figure}

In practical applications, it is preferable to pre-select a small subset of promising candidates based on some compact descriptors before the full-scale similarity calculation\cite{pointnetvlad}. To this end, we propose a low-dimension and $yaw$-invariant retrieval key based on the characteristics of the given contour and its vicinity on the BEV image.

As mentioned in Sec. \ref{sec:relational}, similar CACs have similar anchor CAs. Therefore, the quantified features of anchor CAs should be used in retrieval keys. Among the variables describing a CA (\ref{eq:def_abs_cont}), we use cell count ($n_a$) and the axis length information ($\lambda_{1}, \lambda_{2}$) to create discriminative key dimensions: 
\begin{equation} \label{eq:key_anchor}
\mathbf{K}_{anc}^{ l, s} = \left[ \sqrt{n^{l, s}_{a} \lambda^{l, s}_{1}}, \sqrt{n^{l, s}_{a} \lambda^{l, s}_{2}}, \sqrt{ \Sigma _{i=1}^{s} n^{l, s}_{a}} \right]^{\top}
\end{equation}
where $s$ is the sequence id of a CA at layer $l$. The first two terms generally describe the shape of the CA ellipse. The last term includes the sum over all previous cell counts in the layer, which indicates the relative importance of the current CA in its layer and can penalize similar CAs in a drastically different neighborhood.

The rest key dimensions encode a circular region of interest (RoI) on the BEV image centered at $\mathbf{x}_{c}$ of the CA. To obtain yaw invariance, pixel distance to $\mathbf{x}_{c}$ is the only positional information we use. 
Unlike \cite{scan_context}, we do not create ring keys from occupancy ratio. Instead, we use all the pixels in Euclidean space in the RoI. 
This is because normalization downplays the importance of BEV pixels at faraway places since more pixels are crammed into a distance ring. The absolute number of pixels at every distance level has a wider range, thus can better distinguish different places.

When sorting pixels in the RoI into distance segments, discretization errors can cause significant perturbations. Similar to \cite{glare_2d}, we model the distance of each pixel as a 1D Gaussian distribution and then numerically integrate over the distance segments to get smoothed description values (Fig. \ref{fig:gen_ret_key}). The value of segment $i$ is defined as $\mathbf{K}_{roi}^{ l, s} (i)$ that satisfies: 
\begin{equation} \label{eq:key_vicinity}
\begin{aligned}
\mathbf{K}_{roi}^{ l, s} (i) =& \int_{d_{i-1}}^{d_i} \sum_{\mathbf{p} \in S_{R} }(\operatorname{Lev}(p_z) - l_{b})\cdot \\
&\qquad \qquad \mathcal{N} (x| \| \mathbf{p}_{xy} - \mathbf{x}^{l, s}_{c} \|_{2}, \sigma_{d}) \; dx \\
S_{R} =& \{ \mathbf{p} \; | \; \| \mathbf{p}_{xy} - \mathbf{x}^{l, s}_{c} \|_{2} \le R,\; \operatorname{Lev}(p_z) > l_{b}  \}
\end{aligned}
\end{equation}
where ${d_i}$ is the $i$-th threshold when dividing the distance, $\operatorname{Lev}(h)$ returns the contour level a height $h$ belongs to, $l_b$ is the base level above which a pixel is allowed to contribute to the key, $\sigma_{d}$ is the distribution width parameter for valid BEV pixels, and $R$ is the RoI radius. The physical meaning of the segment value is the number of levels above the base level summed over pixels projected in that distance segment. The complete retrieval key for a given $\mathcal{C}^{l, s}$ is a vector of 
\begin{equation} \label{eq:key_full}
\mathbf{K}^{ l, s} = [w_{1}{\mathbf{K}_{anc}^{ l, s}}^{\top}, \mathbf{K}_{roi}^{ l, s} (1), \cdots, \mathbf{K}_{roi}^{ l, s} (n)]^\top
\end{equation}
where $n$ is the number of distance segments in the RoI, and $w_1$ is a weight parameter since the two parts of the retrieval key have different physical meaning.

Retrieval keys are only created for top CAs in selected layers. A query key only searches the KD-tree of its level. Thanks to the low-dimensional keys, real time is guaranteed in our experiments (Sec. \ref{sec:complexity}) using nanoflann\cite{blanco2014nanoflann}. 

%
%
%

\section{Experiments and Results} 

\subsection{Definition of Evaluation Criteria}
We briefly recapitulate three mainstream protocols that define the ground truth (GT) loop status and determine true or false when given a prediction: 
\begin{enumerate}
	\item \label{crit1} Check the similarity of any assigned pairs. If the spatial distance is smaller than $l_1$, the true loop status is \textit{true}; if the distance is larger than $l_2$ ($l_1<l_2$), the true status is \textit{false}. The prediction result is controlled by a varying similarity threshold. Used by \cite{sg_pr, semantic_sc, rinet}.
	\item \label{crit2} Only evaluate poses with GT loops. Retrieve top-$N$ candidates from all valid past poses. If any candidate is a true neighbor, the retrieval is \textit{true positive} (TP), else \textit{false negative} (FN)\cite{pointnetvlad, disco_lc}. 
	\item \label{crit3} Find the most similar candidate from all valid past poses. When similarity is above a threshold, if the spatial distance to the candidate is smaller than $l_3$, report TP, else \textit{false positive} (FP). When similarity is below the threshold but there exist valid past poses within $l_3$, report FN\cite{lcdnet}.
\end{enumerate}
Our choice is \ref{crit3}) and the reason is twofold. 
Firstly, we aim at global localization so the ability of finding loops among all valid historical scans should be assessed. 
Secondly, the protocol directly requires the best possible candidate which reduces effort in further post-processing. 
In comparison, protocol \ref{crit1}) reports the similarity between assigned pair of proposals, which unnecessarily puts too much emphasis on sub-optimal candidates. 
Protocol \ref{crit2}) cannot evaluate the ability to reject wrong loops (i.e., precision), since they only test for GT loops therefore cannot record FP.

During evaluation, we set $l_3=5m$, and exclude 150 frames right before current frame from loop candidates.


%
\begin{figure*}[htbp]
	
	\centerline{\includegraphics[width=0.9\textwidth]{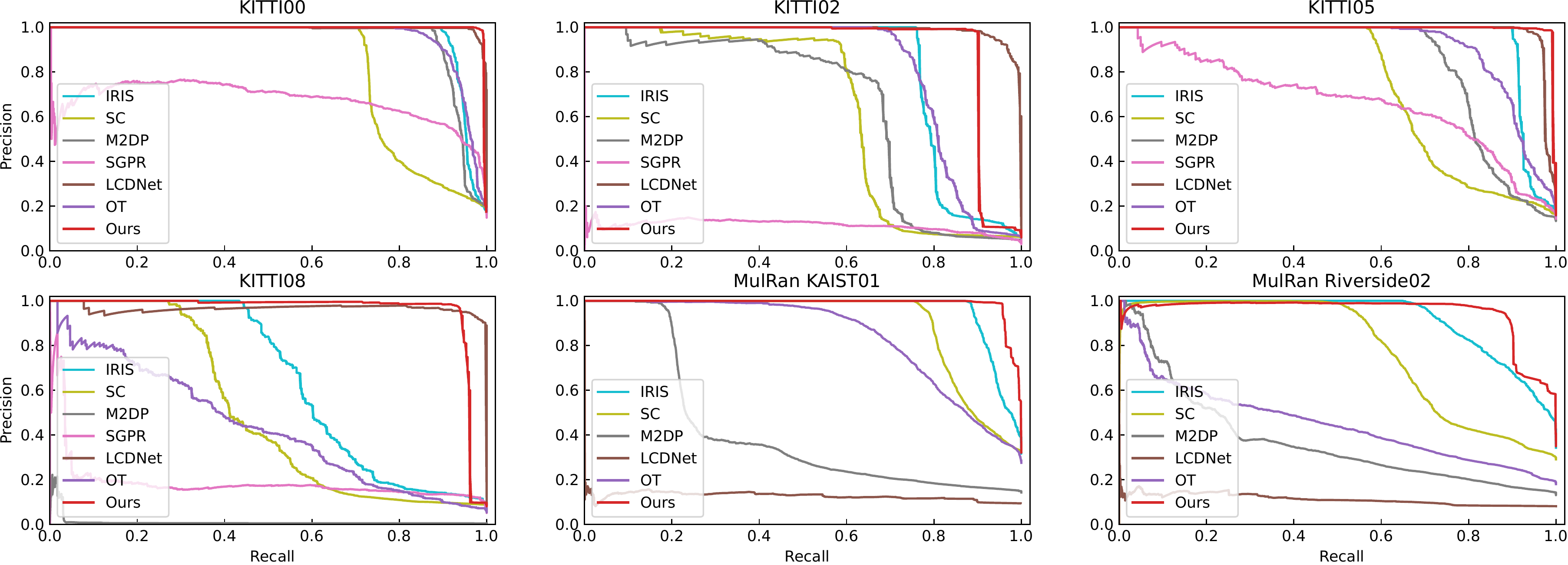}}

	\caption{The precision-recall curves of our method and the six baselines. }
	
	\label{fig:pr}
	
\end{figure*}

\subsection{Evaluation on Public Datasets}
We evaluate on KITTI dataset\cite{kitti_dataset}. Due to GT pose issue \cite{lcdnet}, we use GT poses from SemanticKITTI\cite{semantic_kitti_dataset}.
To show the robustness and versatility of the proposed method, we also test on the MulRan dataset\cite{mulran_dataset}. We only use two different parameters compared with tests on KITTI: BEV slicing heights, mean height difference threshold. The modifications are due to the fact that MulRan's LiDAR has $22.5^{\circ}$ vertical FoV above the horizontal plane ($2^{\circ}$ in KITTI).
Parameters in use can be found in our code repository.

We test on KITTI 00, 02, 05 and 08 (K0x in tables) where loops are abundant, MulRan KAIST01 and Riverside02 (MK01 and MR02 in tables). Six baseline methods are selected for comparison: M2DP\cite{m2dp}, SC\cite{scan_context}, IRIS\cite{lidar_iris}, SGPR\cite{sg_pr}, OT\cite{OverlapTransformer} and LCDNet\cite{lcdnet}, and the last three are learning based (pretrained models in their public code repositories are used for evaluation). MulRan dataset has no semantic labels, so we do not run SGPR on it. To show the generalization ability, both LCDNet and OT directly used pretrained models for real-world tests/other datasets in their own papers\cite{lcdnet, OverlapTransformer}. So we do the same on MulRan. But we note their performance may be better if retrained. 

\begin{table}[tbp]\centering 
	
	\caption{Max F1 score results.} 
	
	\label{tab:maxf1_score}
	
	\begin{threeparttable}
		
		\begin{tabular}{ccccccc}
			
			\toprule
			
			{\textbf{Methods}\tnote{+}} & {\textbf{K00}} & {\textbf{K02}} & {\textbf{K05}} & {\textbf{K08}} & {\textbf{MK01}} & {\textbf{MR02}}\\ 
			
			\midrule
			
			
			M2DP & 0.931 & 0.711 & 0.817 & 0.048 & 0.383 & 0.383 \\
			SC   & 0.827 & 0.718 & 0.727 & 0.504 & 0.862 & 0.697 \\
			IRIS & 0.943 & 0.863 & 0.946 & 0.628 & \underline{0.935} & \underline{0.816} \\
			SGPR(SK)\tnote{*} & 0.705 & 0.208 & 0.658 & 0.272 & -- & -- \\
			OT     & 0.910 & 0.818 & 0.859 & 0.458 & 0.753 & 0.477 \\
			LCDNet & \underline{0.979} & \textbf{0.946} & \underline{0.963} & \textbf{0.954} & 0.215 & 0.191 \\
			Ours   & \textbf{0.988} & \underline{0.939} & \textbf{0.988} & \underline{0.954} & \textbf{0.972} & \textbf{0.913} \\ 
			
			\bottomrule
			
		\end{tabular}
		\begin{tablenotes}\footnotesize
			
			\item[*] SK: use SemanticKITTI semantic labels.
			\item[+] Best scores are emboldened and the second bests are underlined.
		\end{tablenotes}
		
	\end{threeparttable}
	
\end{table}

%

\subsubsection{Precision and Recall}
It is a convention\cite{glare_2d, Steder2011, scan_context, overlapnet } to evaluate topological loop detection with precision-recall curve (PR-curve) and max $F_1$ score. The $F_1$ score ``averages" precision and recall. We report PR-curve in Fig. \ref{fig:pr} and the max $F_1$ score in Tab. \ref{tab:maxf1_score}. According to the results, IRIS is the best handcrafted baseline method, but lacks retrieval for practical usage. LCDNet has exceptional performance on KITTI. Notably, OT\cite{OverlapTransformer} generalizes to MulRan dataset fairly well compared to LCDNet\cite{lcdnet}, while both use pretrained models. Our PR-curve does not perform well on MR02, since it mainly contains trees and open area along a riverbank, and information loss is severe when modeled as distributions. Nevertheless, ours has the best max $F_1$ score.

\subsubsection{Metric Pose Estimation} See Tab. \ref{tab:mpe}. 

\begin{table}[hbtp]\centering 

	\caption{Three DoF MPE for TP loops at max F1 score.} 

	\label{tab:mpe}

	\begin{threeparttable}

		\begin{tabular}{>{\centering\arraybackslash}m{8em}
						m{2em}m{2em}m{2em}m{2em}m{2em}m{2em}} 
			\toprule

			{\textbf{Metrics}} & {\textbf{K00}} & {\textbf{K02}} & {\textbf{K05}} & {\textbf{K08}} & {\textbf{MK01}} & {\textbf{MR02}}\\ 

			\midrule
			Num. TP loops & 795 & 279 & 439 & 323 & 2573 & 2326 \\
			\midrule

			Mean rot. err.(${}^\circ$) & 0.135 & 0.381 & 0.136 & 0.345 & 0.257 & 0.286\\
			RMSE rot.(${}^\circ$)      & 0.189 & 0.537 & 0.195 & 0.471 & 0.480 & 0.436\\
			Mean trans. err.(m)        & 0.120 & 0.730 & 0.132 & 0.202 & 0.245 & 0.288\\
			RMSE trans.(m)             & 0.144 & 0.942 & 0.159 & 0.225 & 0.405 & 0.403\\
			\bottomrule

		\end{tabular}

	\end{threeparttable}

\end{table}

\subsubsection{Computation Complexity}  \label{sec:complexity}
 
Different methods' resource consumption is not directly comparable, as it is implementation dependent. So we only list our time consumption data in Tab. \ref{tab:complexity} for reference. 
The prototype is written in C++ and evaluated
on an Intel Core i7-9750H@2.60GHz laptop CPU with 16GB RAM. No multi-threading is used in code written by ourselves. We run each sequence 3 times then take average. 
Regarding details affecting complexity, we index retrieval keys in 3 levels, each with 6 anchor CAs. Peripheral CAs are top-10 CAs at every level for 4 levels. Each 10-D key retrieves 50 candidates per query. KD-trees are updated every 100 scans, with one level at a time.
We attribute the overall efficiency to the simplicity of elementary operations.

\begin{table}[btp]\centering 

	\caption{Dataset sizes and decomposed average time cost per scan (ms).}

	\label{tab:complexity}

	\begin{threeparttable}

		\begin{tabular}{ccccccc}

			\toprule

			{\textbf{}} & {\textbf{K00}} & {\textbf{K02}} & {\textbf{K05}} & {\textbf{K08}} & {\textbf{MK01}} & {\textbf{MR02}}\\ 

			\midrule

			Valid frames \#  & 4541 & 4661 & 2761 & 4071 & 8034\tnote{*} & 6792\tnote{*} \\

			\midrule
			Gen. contours & 7.9 & 8.7 & 8.5 & 9.0 & 7.9 & 7.1 \\
			Retrieval     & 0.6 & 0.6 & 0.4 & 0.5 & 0.8 & 0.5 \\
			CAC check     & 0.9 & 0.8 & 0.8 & 0.7 & 0.9 & 0.7 \\
			L2 optim.     & 1.2 & 0.4 & 0.9 & 0.7 & 1.7 & 1.0 \\
			Update DB     & 0.1 & 0.1 & 0.1 & 0.1 & 0.1 & 0.1 \\
			\midrule
			Total per scan& 10.6 & 10.6 & 10.7 & 11.0 & 11.4 & 9.5 \\
			\bottomrule

		\end{tabular}
		\begin{tablenotes}\footnotesize
		
			\item[*] Some scans at stationary poses are skipped, since no GT poses.
			
		\end{tablenotes}
	
	\end{threeparttable}

\end{table}

\subsection{Discussion}

\begin{figure}[htbp]
	
	\centerline{\includegraphics[width=0.4\textwidth]{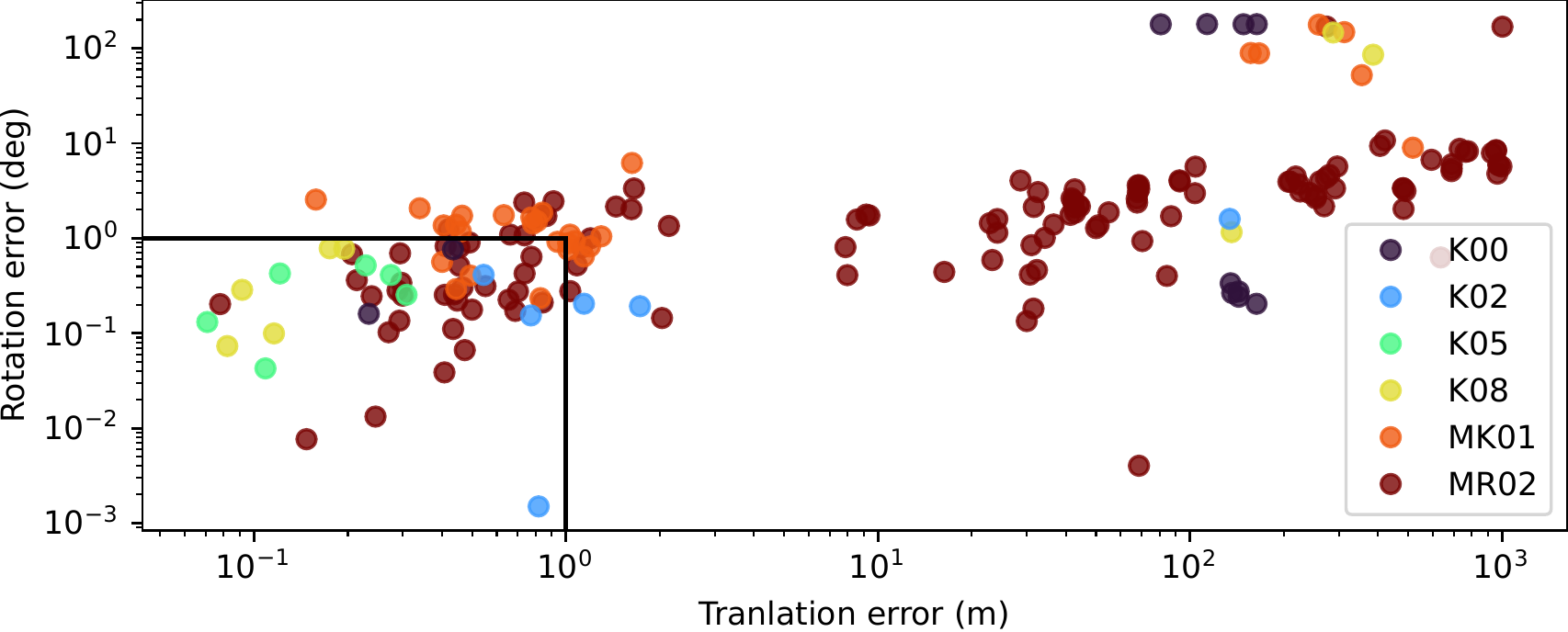}}

	\caption{Metric error distribution of FP predictions at max F1 score. In all six sequences, 52 out of 192 FP predictions ($27\%$) have translation and rotation errors smaller than $1.0m$ and $1.0^{\circ}$ respectively (lower left black rectangle).}
	
	\label{fig:mpe_fp}
	
\end{figure}

\subsubsection{Rethinking evaluation criteria}
In real-world deployment, there is no way to distinguish TP from FP using a single scan. All the positive predictions will go through the same post-processing (e.g., consensus check, ICP). So an interesting question is: are the FP loop closures really that harmful? To answer it, we investigate the MPE error in loop closure events identified as FP by the protocol, and the results are reported in Fig. \ref{fig:mpe_fp}.  
We find that many FPs have metric errors comparable to TP ones, which can act as a good initial guess for ICP\cite{lcdnet}.
This means if we have some mechanisms (e.g., relative pose consensus) to remove extreme outliers, they, though FP by definition, can actually benefit the downstream tasks. Therefore, we believe evaluation criteria that can fully reflect the capability of loop detectors are in demand.


%

\section{Conclusion}

In this paper, we propose a novel loop closure detection method for 3D LiDARs in urban autonomous driving scenario. The method can achieve high precision and recall in terms of topological loop detection performance in real time, and the 3 DoF metric pose estimation results are also accurate without ICP-based 6-DoF fine optimization.


While the BEV representation is an effective simplification, the information loss in the elevation direction is inevitable. The na\"ive contour segmentation also has difficulty with large and natural structures. We may boost the performance by fully utilizing the 3rd dimension of point clouds and relying on more sophisticated segmentation techniques.

\bibliographystyle{IEEEtran}

\bibliography{myrefs}

\begin{thebibliography}{10}
\providecommand{\url}[1]{#1}
\csname url@rmstyle\endcsname
\providecommand{\newblock}{\relax}
\providecommand{\bibinfo}[2]{#2}
\providecommand\BIBentrySTDinterwordspacing{\spaceskip=0pt\relax}
\providecommand\BIBentryALTinterwordstretchfactor{4}
\providecommand\BIBentryALTinterwordspacing{\spaceskip=\fontdimen2\font plus
\BIBentryALTinterwordstretchfactor\fontdimen3\font minus
  \fontdimen4\font\relax}
\providecommand\BIBforeignlanguage[2]{{%
\expandafter\ifx\csname l@#1\endcsname\relax
\typeout{** WARNING: IEEEtran.bst: No hyphenation pattern has been}%
\typeout{** loaded for the language `#1'. Using the pattern for}%
\typeout{** the default language instead.}%
\else
\language=\csname l@#1\endcsname
\fi
#2}}

\bibitem{segmatch}
R.~Dube, D.~Dugas, E.~Stumm, J.~Nieto, R.~Siegwart, and C.~Cadena, ``Segmatch:
  Segment based place recognition in 3d point clouds,'' in \emph{2017 IEEE
  International Conference on Robotics and Automation (ICRA)}, 2017, p.
  5266–5272.

\bibitem{orbslam}
R.~Mur-Artal, J.~M.~M. Montiel, and J.~D. Tardos, ``Orb-slam: A versatile and
  accurate monocular slam system,'' \emph{IEEE Transactions on Robotics},
  vol.~31, no.~5, p. 1147–1163, Oct. 2015.

\bibitem{Lowry_Sunderhauf_Newman_Leonard_Cox_Corke_Milford_2016}
S.~Lowry, N.~Sunderhauf, P.~Newman, J.~J. Leonard, D.~Cox, P.~Corke, and M.~J.
  Milford, ``Visual place recognition: A survey,'' \emph{IEEE Transactions on
  Robotics}, vol.~32, no.~1, p. 1–19, Feb. 2016.

\bibitem{scan_context}
G.~Kim and A.~Kim, ``Scan context: Egocentric spatial descriptor for place
  recognition within 3d point cloud map,'' in \emph{2018 IEEE/RSJ International
  Conference on Intelligent Robots and Systems (IROS)}, 2018, p. 4802–4809.

\bibitem{pointnetvlad}
M.~A. Uy and G.~H. Lee, ``Pointnetvlad: Deep point cloud based retrieval for
  large-scale place recognition,'' in \emph{2018 IEEE/CVF Conference on
  Computer Vision and Pattern Recognition}, 2018, p. 4470–4479.

\bibitem{overlapnet}
X.~Chen, T.~Läbe, A.~Milioto, T.~Röhling, O.~Vysotska, A.~Haag, J.~Behley,
  and C.~Stachniss, ``Overlapnet: Loop closing for lidar-based slam,'' in
  \emph{Robotics: Science and Systems XVI}, 2020.

\bibitem{Bosse2013}
M.~Bosse and R.~Zlot, ``Place recognition using keypoint voting in large 3d
  lidar datasets,'' in \emph{2013 IEEE International Conference on Robotics and
  Automation}, 2013, p. 2677–2684.

\bibitem{segmap_conf}
R.~Dubé, A.~Cramariuc, D.~Dugas, J.~Nieto, R.~Siegwart, and C.~Cadena,
  ``Segmap: 3d segment mapping using data-driven descriptors,'' in
  \emph{Robotics: Science and Systems XIV}, 2018.

\bibitem{scan_context_pp}
G.~Kim, S.~Choi, and A.~Kim, ``Scan context++: Structural place recognition
  robust to rotation and lateral variations in urban environments,'' \emph{IEEE
  Transactions on Robotics}, vol.~38, no.~3, p. 1856–1874, June 2022.

\bibitem{bvmatch}
L.~Luo, S.-y. Cao, B.~Han, H.-l. Shen, and J.~Li, ``Bvmatch: Lidar-based place
  recognition using bird’s-eye view images,'' \emph{IEEE Robotics and
  Automation Letters}, vol.~6, no.~3, p. 6076–6083, July 2021.

\bibitem{disco_lc}
X.~Xu, H.~Yin, Z.~Chen, Y.~Li, Y.~Wang, and R.~Xiong, ``Disco: Differentiable
  scan context with orientation,'' \emph{IEEE Robotics and Automation Letters},
  vol.~6, no.~2, p. 2791–2798, Apr. 2021.

\bibitem{lidar_iris}
Y.~Wang, Z.~Sun, C.-Z. Xu, S.~E. Sarma, J.~Yang, and H.~Kong, ``Lidar iris for
  loop-closure detection,'' in \emph{2020 IEEE/RSJ International Conference on
  Intelligent Robots and Systems (IROS)}, 2020, p. 5769–5775.

\bibitem{spi_lc}
Y.~Li and H.~Li, ``Lidar-based initial global localization using
  two-dimensional (2d) submap projection image (spi),'' in \emph{2021 IEEE
  International Conference on Robotics and Automation (ICRA)}, 2021, p.
  5063–5068.

\bibitem{lcdnet}
D.~Cattaneo, M.~Vaghi, and A.~Valada, ``Lcdnet: Deep loop closure detection and
  point cloud registration for lidar slam,'' \emph{IEEE Transactions on
  Robotics}, p. 1–20, 2022.

\bibitem{m2dp}
L.~He, X.~Wang, and H.~Zhang, ``M2dp: A novel 3d point cloud descriptor and its
  application in loop closure detection,'' in \emph{2016 IEEE/RSJ International
  Conference on Intelligent Robots and Systems (IROS)}, 2016, p. 231–237.

\bibitem{glare_2d}
M.~Himstedt, J.~Frost, S.~Hellbach, H.-J. Bohme, and E.~Maehle, ``Large scale
  place recognition in 2d lidar scans using geometrical landmark relations,''
  in \emph{2014 IEEE/RSJ International Conference on Intelligent Robots and
  Systems}, Chicago, IL, USA, 2014, p. 5030–5035.

\bibitem{OverlapTransformer}
J.~Ma, J.~Zhang, J.~Xu, R.~Ai, W.~Gu, and X.~Chen, ``Overlaptransformer: An
  efficient and yaw-angle-invariant transformer network for lidar-based place
  recognition,'' \emph{IEEE Robotics and Automation Letters}, vol.~7, no.~3, p.
  6958–6965, July 2022.

\bibitem{Blanco2013}
J.-L. Blanco, J.~González-Jiménez, and J.-A. Fernández-Madrigal, ``A robust,
  multi-hypothesis approach to matching occupancy grid maps,'' \emph{Robotica},
  vol.~31, no.~5, p. 687–701, Aug. 2013.

\bibitem{segmap_j}
R.~Dubé, A.~Cramariuc, D.~Dugas, H.~Sommer, M.~Dymczyk, J.~Nieto, R.~Siegwart,
  and C.~Cadena, ``Segmap: Segment-based mapping and localization using
  data-driven descriptors,'' \emph{The International Journal of Robotics
  Research}, vol.~39, no. 2–3, p. 339–355, Mar. 2020.

\bibitem{Dube2018}
R.~Dube, M.~G. Gollub, H.~Sommer, I.~Gilitschenski, R.~Siegwart, C.~Cadena, and
  J.~Nieto, ``Incremental-segment-based localization in 3-d point clouds,''
  \emph{IEEE Robotics and Automation Letters}, vol.~3, no.~3, p. 1832–1839,
  July 2018.

\bibitem{sg_pr}
X.~Kong, X.~Yang, G.~Zhai, X.~Zhao, X.~Zeng, M.~Wang, Y.~Liu, W.~Li, and
  F.~Wen, ``Semantic graph based place recognition for 3d point clouds,'' in
  \emph{2020 IEEE/RSJ International Conference on Intelligent Robots and
  Systems (IROS)}, Las Vegas, NV, USA, 2020, p. 8216–8223.

\bibitem{gosmatch}
Y.~Zhu, Y.~Ma, L.~Chen, C.~Liu, M.~Ye, and L.~Li, ``Gosmatch:
  Graph-of-semantics matching for detecting loop closures in 3d lidar data,''
  in \emph{2020 IEEE/RSJ International Conference on Intelligent Robots and
  Systems (IROS)}, Las Vegas, NV, USA, 2020, p. 5151–5157.

\bibitem{Fan_He_Tan_2020}
Y.~Fan, Y.~He, and U.-X. Tan, ``Seed: A segmentation-based egocentric 3d point
  cloud descriptor for loop closure detection,'' in \emph{2020 IEEE/RSJ
  International Conference on Intelligent Robots and Systems (IROS)}, Las
  Vegas, NV, USA, 2020, p. 5158–5163.

\bibitem{boxgraph}
G.~Pramatarov, D.~De~Martini, M.~Gadd, and P.~Newman, ``Boxgraph: Semantic
  place recognition and pose estimation from 3d lidar,'' in \emph{2022 IEEE/RSJ
  International Conference on Intelligent Robots and Systems (IROS)}, Kyoto,
  Japan, 2022, p. 7004–7011.

\bibitem{locus_lc}
K.~Vidanapathirana, P.~Moghadam, B.~Harwood, M.~Zhao, S.~Sridharan, and
  C.~Fookes, ``Locus: Lidar-based place recognition using spatiotemporal
  higher-order pooling,'' in \emph{2021 IEEE International Conference on
  Robotics and Automation (ICRA)}, Xi’an, China, 2021, p. 5075–5081.

\bibitem{gmmreg}
B.~Jian and B.~C. Vemuri, ``Robust point set registration using gaussian
  mixture models,'' \emph{IEEE Transactions on Pattern Analysis and Machine
  Intelligence}, vol.~33, no.~8, p. 1633–1645, Aug. 2011.

\bibitem{ceres}
\BIBentryALTinterwordspacing
S.~Agarwal, K.~Mierle, and T.~C.~S. Team, ``{Ceres Solver},'' 3 2022. [Online].
  Available: \url{https://github.com/ceres-solver/ceres-solver}
\BIBentrySTDinterwordspacing

\bibitem{l2_3d_ndt}
T.~Stoyanov, M.~Magnusson, and A.~J. Lilienthal, ``Point set registration
  through minimization of the l2 distance between 3d-ndt models,'' in
  \emph{2012 IEEE International Conference on Robotics and Automation}, St
  Paul, MN, USA, 2012, p. 5196–5201.

\bibitem{blanco2014nanoflann}
J.~L. Blanco and P.~K. Rai, ``nanoflann: a {C}++ header-only fork of {FLANN}, a
  library for nearest neighbor ({NN}) with kd-trees,''
  \url{https://github.com/jlblancoc/nanoflann}, 2014.

\bibitem{semantic_sc}
L.~Li, X.~Kong, X.~Zhao, T.~Huang, W.~Li, F.~Wen, H.~Zhang, and Y.~Liu, ``Ssc:
  Semantic scan context for large-scale place recognition,'' in \emph{2021
  IEEE/RSJ International Conference on Intelligent Robots and Systems (IROS)},
  Prague, Czech Republic, 2021, p. 2092–2099.

\bibitem{rinet}
------, ``Rinet: Efficient 3d lidar-based place recognition using rotation
  invariant neural network,'' \emph{IEEE Robotics and Automation Letters},
  vol.~7, no.~2, p. 4321–4328, Apr. 2022.

\bibitem{kitti_dataset}
A.~Geiger, P.~Lenz, C.~Stiller, and R.~Urtasun, ``Vision meets robotics: The
  kitti dataset,'' \emph{The International Journal of Robotics Research},
  vol.~32, no.~11, p. 1231–1237, Sept. 2013.

\bibitem{semantic_kitti_dataset}
J.~Behley, M.~Garbade, A.~Milioto, J.~Quenzel, S.~Behnke, C.~Stachniss, and
  J.~Gall, ``Semantickitti: A dataset for semantic scene understanding of lidar
  sequences,'' \emph{arXiv:1904.01416}, Aug. 2019.

\bibitem{mulran_dataset}
G.~Kim, Y.~S. Park, Y.~Cho, J.~Jeong, and A.~Kim, ``Mulran: Multimodal range
  dataset for urban place recognition,'' in \emph{2020 IEEE International
  Conference on Robotics and Automation (ICRA)}, Paris, France, 2020, p.
  6246–6253.

\bibitem{Steder2011}
B.~Steder, M.~Ruhnke, S.~Grzonka, and W.~Burgard, ``Place recognition in 3d
  scans using a combination of bag of words and point feature based relative
  pose estimation,'' in \emph{2011 IEEE/RSJ International Conference on
  Intelligent Robots and Systems}, 2011, p. 1249–1255.

\end{thebibliography}

\end{document}